\pdfoutput=1

\documentclass[11pt]{article}

\usepackage[preprint]{coling}

\usepackage{times}
\usepackage{latexsym}
\usepackage{tikz}
\usepackage{varwidth}
\usepackage{tcolorbox}
\usetikzlibrary{positioning}
\usepackage{amsmath}

\usepackage{array}
\usepackage{tabularx}
\usepackage{authblk}

\usepackage[T1]{fontenc}

\usepackage[utf8]{inputenc}

\usepackage{microtype}

\usepackage{inconsolata}

\usepackage{graphicx}

\usepackage[utf8]{inputenc} %
\usepackage[T1]{fontenc}    %
\usepackage{hyperref}       %
\usepackage{url}            %
\usepackage{booktabs}       %
\usepackage{multirow}       %
\usepackage{rotating}
\usepackage{makecell}
\usepackage{adjustbox}
\usepackage{amsfonts}       %
\usepackage{nicefrac}       %
\usepackage{microtype}      %
\usepackage{xcolor}         %

\usepackage{graphicx}
\usepackage{subcaption}
\usepackage{wrapfig}
\usepackage{adjustbox}
\usepackage{amsthm}
\usepackage{amsmath}
\usepackage{amssymb}
\usepackage{amsfonts}
\usepackage{algorithm2e}
\usepackage{mathtools}
\usepackage{setspace}
\usepackage{pbalance}

\usepackage[normalem]{ulem}
\usepackage{bm}
\usepackage{bbm}
\usepackage{enumitem}
\usepackage{tabularx}

\newcommand{\TODO}[2][\relax]{%
  \textcolor{red}{TODO\ifx#1\relax\else\ (#1)\fi: #2}%
}

\theoremstyle{definition}

\RestyleAlgo{ruled}
\SetKwInOut{Input}{input}\SetKwInOut{Output}{output}

\newcolumntype{R}[2]{%
    >{\adjustbox{angle=#1,lap=\width-(#2)}\bgroup}%
    l%
    <{\egroup}%
}

\newcommand{\figsize}{0.195\textwidth}
\newcommand{\R}{\mathbbm{R}}

\title{On Evaluating LLMs' Capabilities as Functional Approximators: \\ A Bayesian Perspective}

\author[1]{Shoaib Ahmed Siddiqui\thanks{Authors contributed equally}}
\author[1]{Yanzhi Chen\protect\footnotemark[1]}
\author[1]{Juyeon Heo} 
\author[2]{Menglin Xia} 
\author[1,3]{Adrian Weller} 

\affil[1]{University of Cambridge}
\affil[2]{Microsoft}
\affil[3]{The Alan Turing Institute}

\begin{document}
\maketitle
\begin{abstract}
Recent works have successfully applied Large Language Models (LLMs) to function modeling tasks. However, the reasons behind this success remain unclear. In this work, we propose a new evaluation framework to comprehensively assess LLMs' function modeling abilities. By adopting a Bayesian perspective of function modeling, we discover that LLMs are relatively weak in understanding patterns in raw data, but excel at utilizing prior knowledge about the domain to develop a strong understanding of the underlying function. Our findings offer new insights about the strengths and limitations of LLMs in the context of function modeling.
\end{abstract}

\section{Introduction}

Large Language Models (LLMs) have revolutionized domains that can be formulated in a simple text-in-text-out format~\citep{raffel2020exploring,brown2020gpt3}.
This includes a wide array of tasks from a helpful chatbot~\citep{achiam2023gpt4}, code assistant~\citep{roziere2023codellama}, math theorem provers~\citep{trinh2024alphageometry}, to an automated scientist~\citep{lu2024aiscientist}.

Given the well-established performance characteristics of LLMs in a wide range of reasoning tasks~\citep{achiam2023gpt4,bubeck2023sparks,si2024llmnlpresearch,lu2024aiscientist}, there has been growing interest in exploring their application on real-world prediction tasks i.e., as a regression system~\citep{codeaspolicies2022,gpt4nas,roberts2023gpt4geo,yu2023temporal,xiao2024verbalized}. These approaches typically convert numerical data into tokens that can be processed by an LLM, which then predicts numerical targets for a query point conditioned on the task description and the provided in-context examples.

\begin{figure}[t!]
    \centering
    \begin{tikzpicture}
        \draw[fill=cyan] (-1.5,-0.5) -- (-1.7,0.2) -- (-0.1,0.5) -- (0.0,0.2) -- cycle;
        \draw[fill=gray!70] (-1.5,-0.5) circle (0.3);
    
        \filldraw[black] (0.4,0.5) circle (2pt);
        \filldraw[black] (1.0,0.65) circle (2pt);
        \filldraw[black] (1.6,0.8) circle (2pt);
    
        \draw[densely dotted, thick, red, -{>[scale=1.5]}] (0.4,0.5) -- (5.0,1.7);
        \draw[thick, red] (5.1,1.725) circle (2pt);
        \node[align=center] at (4.3,2.2) {\small  w/o  \small domain knowledge};
    
    \draw[thick, blue, -{>[scale=1.5]}] (0.4,0.5) .. controls (1.0,0.8) and (1.6,0.85) .. (2.2,0.9) .. controls (2.8,0.95) and (3.2,1.0) .. (5.0,0.72);
    \draw[thick, blue] (5.1,0.725) circle (2pt);
    \node[align=center] at (4.3, 0.15) {\small w/ \small domain knowledge};
    
        \node at (5.1,1.2) {?};
        
    \end{tikzpicture}
    \caption{\textbf{A motivating example}. When making predictions, a model focusing only on raw data may interpret the underlying function as a linear one. However, when domain information is specified (i.e., the trajectory of a cannonball), the model can take into account physical laws for more accurate modeling of the trajectory. Given the vast amount of knowledge gathered during pretraining, LLMs can integrate domain knowledge they possess to generate more accurate predictions. We are interested in separately evaluating LLMs' ability of understanding raw data patterns and the ability of utilizing domain knowledge in function modelling tasks. }
    \label{fig:motivation}
    \vspace{-5mm}
\end{figure}
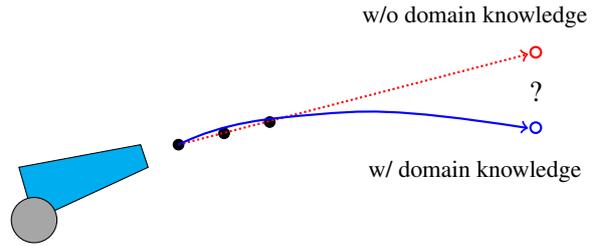

Prominent examples of using LLMs for function approximation include predicting velocities in robotics~\citep{codeaspolicies2022}, predicting the accuracy of models with different configurations for architecture search~\citep{gpt4nas}, directly predicting elevation given geospatial coordinates~\citep{roberts2023gpt4geo}, or even direct forecasting of time-series data~\citep{yu2023temporal,requeima2024llmprocesses}. Compared to conventional machine learning approaches, one immediate advantage of this LLM-based prediction paradigm is the natural ability to condition the model output on arbitrary side information (including detailed task description) provided in the form of natural language. This side information enables the language model to capture the underlying function accurately based on the rich prior knowledge it acquired in pretraining.

Despite the reported successes of LLMs in function modeling tasks, the underlying reasons for this performance remain poorly understood. Our work focuses on two key questions: (i) \textit{can LLMs really comprehend patterns in raw data?}, and (ii) \textit{to what extent can LLMs integrate and utilize domain-specific knowledge in function modeling?}
As illustrated in Fig.~\ref{fig:motivation}, domain knowledge can shape a strong prior for the underlying function and significantly affect prediction. Addressing these questions is crucial for the reliable and effective use of LLMs in real-world prediction tasks.

In order to help shed light on these questions, we present a novel evaluation framework for comprehensively assessing LLMs' function modeling capabilities. Our framework is inspired by principles from Bayesian machine learning, where we separate LLMs' function modeling abilities into two components: their ability to recognize patterns in raw data $\mathcal{D}$ (corresponding to the quality of the likelihood $p(\mathcal{D}|f)$, where $f$ is the function to model) and their ability to incorporate domain knowledge possesses by the LLM (corresponding to the quality of the posterior $p(f|\mathcal{D}) \propto p(\mathcal{D}|f)p(f)$, with $p(f)$ the prior shaped by LLMs' domain knowledge). With this framework, we are able to pinpoint the strengths and weaknesses of state-of-the-art LLMs in function modeling tasks. Our contributions are:

\begin{itemize}[leftmargin=*]
    \item We propose a novel evaluation framework to comprehensively assess LLMs' function modeling capabilities, disentangling their ability to understand data patterns from their ability to incorporate prior domain knowledge.
    \item We evaluate the performance of state-of-the-art LLMs (e.g., GPT-4) on both synthetic and real-world prediction tasks, highlighting their strength and weaknesses as function approximators.
\end{itemize}

\section{Background and Related Work}

\subsection{Large language models}

Large Language Models (LLMs) are probabilistic models trained to predict the probability distribution over the next token conditioned on previous tokens~\citep{radford2018gpt} i.e.,
\begin{equation}
    \textsc{LLM}_{\theta}(w_1, ..., w_t; \theta) := p(w_{t+1} | w_1, ..., w_t)
\end{equation}
Outputs from the model are generated via auto-regressive sampling of the next token $w_{t+1} \sim p_{\theta}(w_{t+1}|w_1, ..., w_t)$ conditioned on all the previous tokens~\citep{radford2018gpt}.
After training on massive datasets spanning trillions of tokens, the model acquires a vast amount of knowledge and reasoning capabilities~\citep{bubeck2023sparks,dubey2024llama}.
This knowledge can be leveraged by the model for better function modeling e.g., by taking physical constraints into account when modeling a physical phenomenon (see Fig.~\ref{fig:motivation}).

\subsection{LLMs as functional predictors}
Given prompt $t$ describing the prediction task, systems that use LLMs to make prediction can be mathematically described as follows:
\begin{equation}
\hat{y}(x;t, \mathcal{D}) \vcentcolon = {\tt{EXTRACT}}(\mathop{\arg\max}_s p(s|x ,t, \mathcal{D}))
\label{eq:llm-as-predictor}
\end{equation}
where $p$ is modeled via the language model $\textsc{LLM}_{\theta}$, $x$ is the query datapoint, $t$ is the task description, $\mathcal{D}$ is the data used for in-context learning, and $\tt{EXTRACT}(\cdot)$ extracts the prediction $\hat{y}$ from the generated sequence $s$
\footnote{A regular expression parser can be sufficient in most cases, which requires the outputs of LLM to satisfy a certain format. This can be enforced by specifying it in the task description $t$.}.
Note that the in-context learning data can also be empty i.e., $\mathcal{D} = \varnothing$.

Prior work applying LLMs in different function modeling tasks can be seen as different instantiations of the same prediction framework.
\citet{codeaspolicies2022} used LLMs to convert context-dependent terms such as `more` or `less` to exact velocity values. This can be seen as modeling the function between the terms and the velocity.
\citet{gpt4nas} used GPT-4 to predict the accuracy of models with different configurations as a more efficient architecture search scheme, which can be seen as modeling the function behind hyperparameter configurations and accuracy.
GPT4Geo~\citep{roberts2023gpt4geo} attempted to leverage GPT-4 to directly predict elevation given geospatial coordinates, thereby also modeling a real-world function. \citet{yu2023temporal} evaluated the capabilities of LLMs to directly forecast financial time series. 
Similarly, \citet{gruver2023zeroshottimeseriesforecaster} used LLMs to directly forecast time-series values while demonstrating their capability to model distributions. \citet{requeima2024llmprocesses} introduced the idea of LLM Processes and applied LLM to a number of different time-series prediction tasks by conditioning on additional side information. \citet{qin2023chatgpt} conducted a comprehensive empirical assessment of GPT-4's capabilities across a spectrum of arithmetic reasoning tasks. 
\citet{bubeck2023sparks} also presented an example of function modeling tasks by evaluating the capabilities of GPT-4 in solving math riddles.

Unlike prior work that applied language models to novel function modeling tasks, we instead attempt to understand the reasons for their success by using a Bayesian evaluation framework, where we disentangle the language model's function modeling capabilities into two fundamental aspects.

\section{A Bayesian Evaluation Framework}
\label{sec:methods}

\begin{figure*}[t]
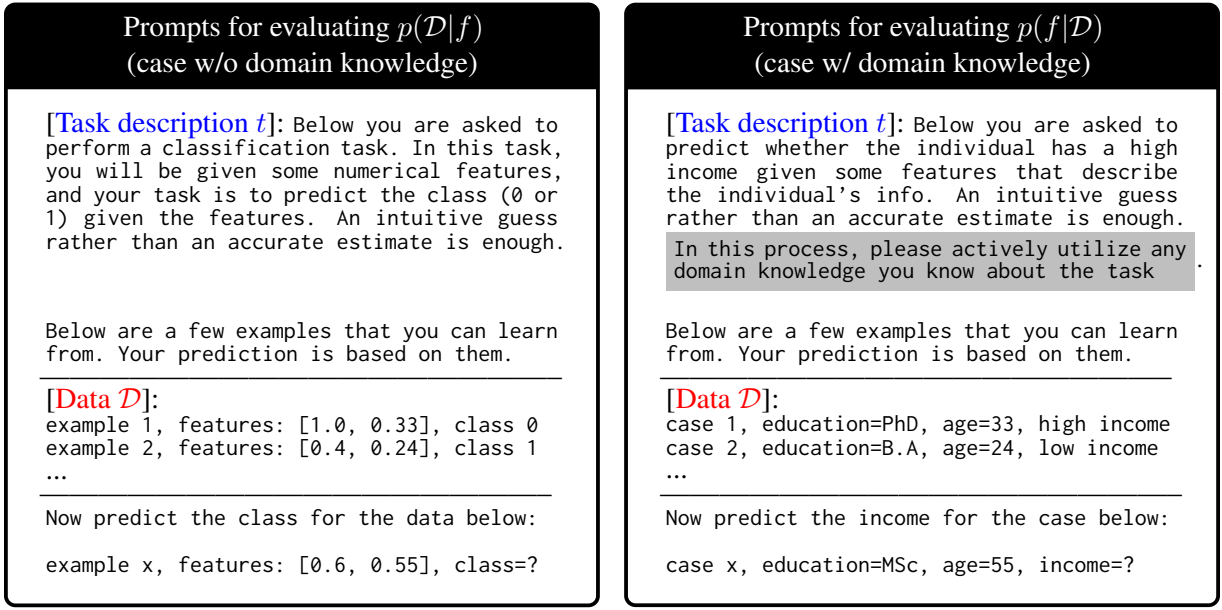

    \label{fig:eval-prompts}
    \centering
    \begin{minipage}{0.49\textwidth}
        \begin{tcolorbox}[colback=white, colframe=black, title={\centering Prompts for evaluating $p(\mathcal{D}|f)$ \\ (case w/o domain knowledge)}]
        \begin{spacing}{0.65}
        [\textcolor{blue}{Task description $t$}]: {\tt \small Below you are asked to perform a classification task. In this task, you will be given some numerical features, and your task is to predict the class (0 or 1) given the features. An intuitive guess rather than an accurate estimate is enough. } \\\\

        \vspace{-0.06cm}

        {\tt \small Below are a few examples that you can learn from. Your prediction is based on them.   } \\
        ----------------------------------------------------- 
        [\textcolor{red}{Data $\mathcal{D}$}]: 

        {\tt \small example 1, features: [1.0, 0.33], class 0 \\
        example 2, features: [0.4, 0.24], class 1} \\
        ... \\
        ----------------------------------------------------- 
        
        {\tt \small Now predict the class for the data below:} \\\\
        {\tt \small example x, features: [0.6, 0.55], class=?} 
        \vspace{-0.25cm}
        \end{spacing}
        \end{tcolorbox}
    \end{minipage} 
    \hfill %
    \begin{minipage}{0.49\textwidth}
        \begin{tcolorbox}[colback=white, colframe=black, title={\centering Prompts for evaluating $p(f|\mathcal{D})$ \\ (case w/ domain knowledge)}]
        \begin{spacing}{0.65}
          [\textcolor{blue}{Task description $t$}]: {\tt \small Below you are asked to predict whether the individual has a high income given some features that describe the individual's info. An intuitive guess rather than an accurate estimate is enough. \colorbox{lightgray}{\parbox{\linewidth}{In this process, please actively utilize any domain knowledge you know about the task}}.} \\\\

        {\tt \small Below are a few examples that you can learn from. Your prediction is based on them. } \\
        ----------------------------------------------------- 

        [\textcolor{red}{Data $\mathcal{D}$}]: 

        {\tt \small case 1, education=PhD, age=33, high income \\
        case 2, education=B.A, age=24, low income} \\
        ... \\
        ----------------------------------------------------- 
        {\tt \small Now predict the income for the case below:} \\\\
        {\tt \small case x, education=MSc, age=55, income=?} 
        \vspace{-0.25cm}
        \end{spacing}
        \end{tcolorbox}
    \end{minipage}
    \caption{Example prompt configurations for evaluating the quality of the likelihood $p(\mathcal{D}|f)$ and the posterior $p(f|\mathcal{D})$ encoded by the LLM in a function modeling task. When evaluating the posterior, a prompt (highlighted in color \colorbox{lightgray}{gray}) is used to explicitly encourage the LLM to make use of domain knowledge regarding the task. }
    \label{fig:eval-prompts}
\end{figure*}

\paragraph{Function modeling as Bayesian inference.} We begin by framing the task of real-world function modeling as performing Bayesian inference in functional space~\cite{ghahramani2015probabilistic}:
\begin{equation}
    p(f|\mathcal{D}) \propto p(\mathcal{D}|f)p(f)
    \label{formula:bayesian-framework}
\end{equation}
where $\mathcal{D}$ is the data, $p(\mathcal{D}|f)$ is the likelihood function that measures to what extent a function $f$ matches with the data, and $p(f)$ is the prior over $f$.
Both $p(\mathcal{D}|f)$ and $p(f)$ are important for accurate modeling of the function. For example, a linear function $f_{\text{linear}}$ that accurately describes the trajectory of a ball in Fig.~\ref{fig:motivation} may attain a high value of likelihood $p(\mathcal{D}|f)$, yet a good prior $p(f)$ based on physics would identify that a quadratic function $f_{\text{quadratic}}$ is indeed more plausible, given context information about the task.

Motivated by this Bayesian view of function modelling, we propose to factorize LLMs' function modeling capabilities into two key aspects:
\begin{itemize}[leftmargin=*]
    \item The ability to understand the patterns present in \emph{raw} data. This corresponds to the quality of the likelihood function $p(\mathcal{D}|f)$ in Eq.~\ref{formula:bayesian-framework}.
    \item The ability to incorporate domain knowledge in order to better estimate the underlying function. This corresponds to the quality of the posterior $p(f|\mathcal{D})$ over the function in Eq.~\ref{formula:bayesian-framework}, with $p(f)$ being the prior shaped by the domain knowledge acquired by the LLM during pretraining.
\end{itemize}
The prior $p(f)$ itself can be viewed as a posterior over $f$ after seeing the massive data  $\mathcal{D}_{\text{web}}$ on internet during LLM pretraining: $p(f) = p(f|\mathcal{D}_{\text{web}})$.

\paragraph{Evaluation objectives.} We are interested in separately evaluating the aforementioned two capabilities of language models in function modeling tasks. This separate evaluation, supported by our techniques detailed below, enable us to understand different aspects of the language model's capabilities. Here, we assess them by the prediction accuracy ${\tt{ACC}}$ of the language model on the test set $\mathcal{D}_{\text{test}}$:
\begin{equation}
    {\tt{ACC}}(\mathcal{D}, t) = \mathbb{E}_{(x, y) \sim \mathcal{D}_{\text{test}}}\big[\mathbf{1}[\hat{y}(x;\mathcal{D}, t) = y]\big]
\end{equation}
By carefully specifying the data $\mathcal{D}$ and the prompt $t$, we can either utilize or disregard domain knowledge provided by the language model during prediction, leading to the isolated evaluation of the aforementioned two capabilities of the model.

\subsection{Evaluating the ability to understand raw data patterns}

In this section, we develop techniques for evaluating the quality of the ability of LLM to understand the patterns in raw data. The key to this evaluation is to remove the influence of the prior $p(f)$, where we ensure that no domain knowledge can be utilized by the LLM in its prediction. We realize this through two important operations applied to the prompts: $\tt{NUMERIZE}(\cdot)$ and $\tt{DECONTEXTUALIZE}(\cdot)$. \\

\noindent \textbf{Numerizing data}. We first remove any information about domain by turning each data $x$ in the original data $\mathcal{D}$, which is potentially in verbal form, into purely numerical values:
\begin{equation*}
    x \leftarrow \tt{NUMERIZE}(x), \quad \forall x \in \mathcal{D}
    \label{eq:numerize}
\end{equation*}
where $\tt{NUMERIZE}: \mathcal{S} \to \R^d$ is an operation that to map a sentence $s \in \mathcal{S}$ from the sentence space $\mathcal{S}$ to a real-valued vector. For example, the sentence {`\{education=PhD, age=33\}'} describing the features of an individual will be transformed to a datum $z=\{1.0, 0.33\}$. Here, the values $z_i$ are normalized, so that $z_i \in [0, 1]$. Normalization is introduced to avoid a LLM from inferring the semantic meaning of features according to their scale, range, and distribution\footnote{As an example, consider $z_1$ to encode the age of individuals. After observing a large number of values of $z_1$ in $\mathcal{D}$, a powerful LLM may infer that this feature corresponds to the age from the distribution and the range of the feature values.}.
Note that a similar operation is usually applied before feeding data into classical machine learning approaches due to their inherent inability to condition on arbitrary text.

\paragraph{Decontextualizing task description.} Another operation is to remove any information about the domain or the context from the task description $t$:
\begin{equation*}
    t \leftarrow {\tt{DECONTEXTUALIZE}}(t)
     \label{eq:decontexualize}
\end{equation*}
where $\tt{DECONTEXTUALIZE}: \mathcal{S} \to \mathcal{S}$ is an operation to rewrite the task description. For example, consider the original task description in the ball trajectory prediction example where $t$ can be something like {`we would like to predict the trajectory of a ball given its past trajectory'}. The $\tt{DECONTEXTUALIZE}$ function rewrites this task description as {`this is a regression task where we predict y from x given some training data'}, stripping away any possible cues to leak domain information.

Fig.~\ref{fig:eval-prompts} provides an example of the prompt we use to evaluate this ability.

\subsection{Evaluating the ability to incorporate domain knowledge}
In this section, we focus on techniques for evaluating the ability of LLM to incorporate domain knowledge in function modeling tasks. Unlike the previous case where we remove the impact of prior, here we aim to emphasize the influence of the prior $p(f)$, which represents the domain knowledge the LLM holds. We achieve this by two operations applied to the prompts: $\tt{VERBALIZE}(\cdot)$ and $\tt{CONTEXUALIZE}(\cdot)$.

\paragraph{Verbalizing data.} Our first operation corresponds to rewriting each data $x$ in the original dataset $\mathcal{D}$ by `verbalizing' it:
\begin{equation*}
        x \leftarrow \tt{VERBALIZE}(x), \quad \forall x \in \mathcal{D}
    \label{eq:verbalize}
\end{equation*}
where $\tt{VERBALIZE}: \mathcal{S} \to \mathcal{S}$: is a function transforms all numerical features in the original data to its natural language-based representation. During this process, the semantic meaning of each feature will also be clearly indicated if they are not specified in the original data. For example, the sentence $s$=`{$\{1, \text{'married'}, 27\}$' will be rewritten as $s'$=`\{\text{Gender=Female}, \text{Marriage status='married'}, \text{Age=}27\}' to better convey the context of the data. This operation can be seen as the reverse operation of the previous $\tt{NUMERIZE}(\cdot)$ operation.

\paragraph{Amplifying the impact of prior.} Our second operation is to add context information to the task description as well as explicitly prompting the LLM to actively make use of any prior knowledge. Specifically, we rewrite the task description $t$ as:
\begin{equation*}
    t \leftarrow ({\tt{CONTEXUALIZE}}(t), h)
\end{equation*}
where $\tt{CONTEXUALIZE(\cdot): \mathcal{S} \to \mathcal{S}}$ is a function that adds context information of the task (e.g. domain, explanation of features, data source, etc.) and $h \in \mathcal{S}$ is an additional `hinting' prompt concatenated to the contexualized prompt. This hinting prompt $h$ is to trigger LLM to explicitly rely on domain knowledge it possesses when making prediction. One instance of $h$ is `during the process, please actively make use of any domain knowledge or prior information you know about [keywords] and incorporate it with the patterns you see from the data.', with [keywords] being the name of the domain e.g., law, physics, finance, medicine, etc.

Fig.~\ref{fig:eval-prompts} provides an example of the prompt we use to evaluate this ability.

\begin{figure*}
    \centering
    \begin{subfigure}[b]{\figsize}
        \centering
        \includegraphics[width=\textwidth]{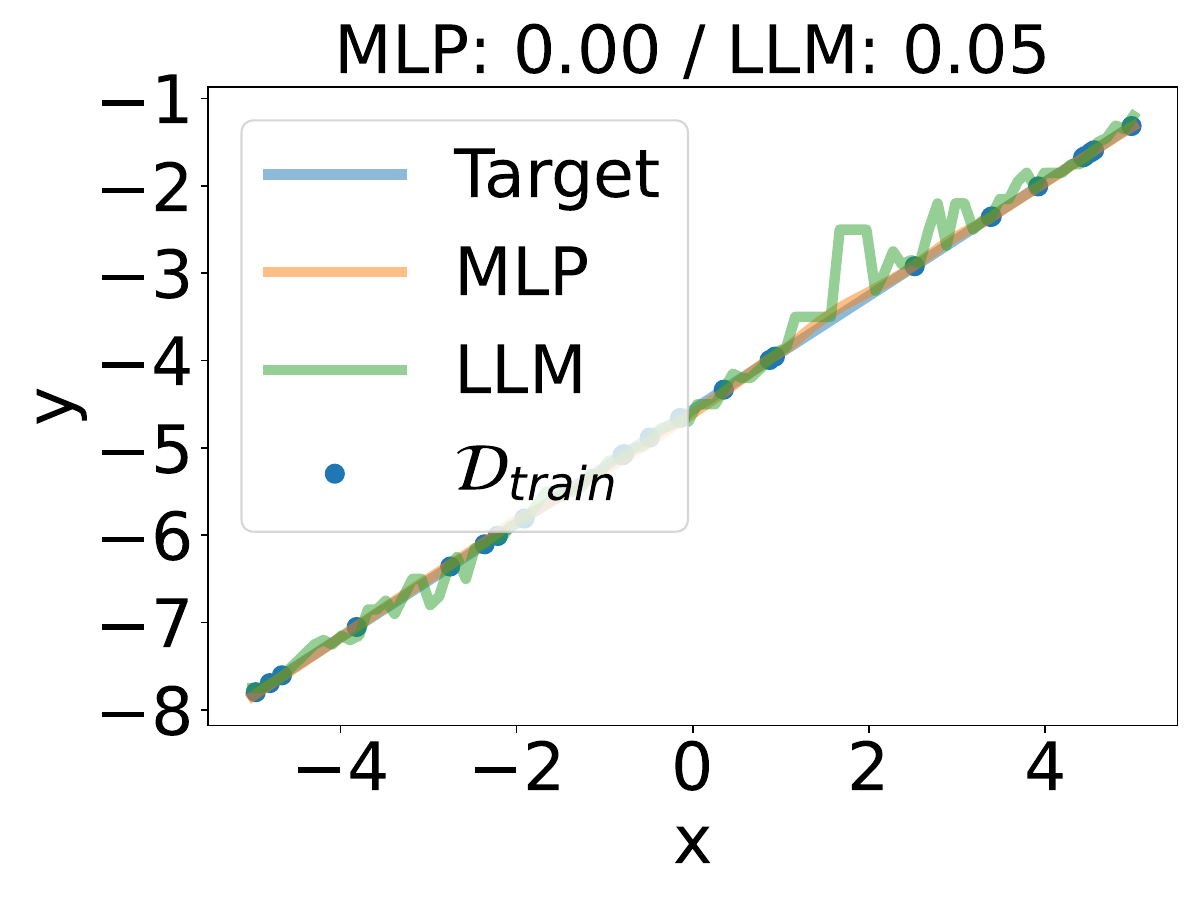}
        \caption{Linear}
    \end{subfigure}
    \hfill
    \begin{subfigure}[b]{\figsize}
        \centering
        \includegraphics[width=\textwidth]{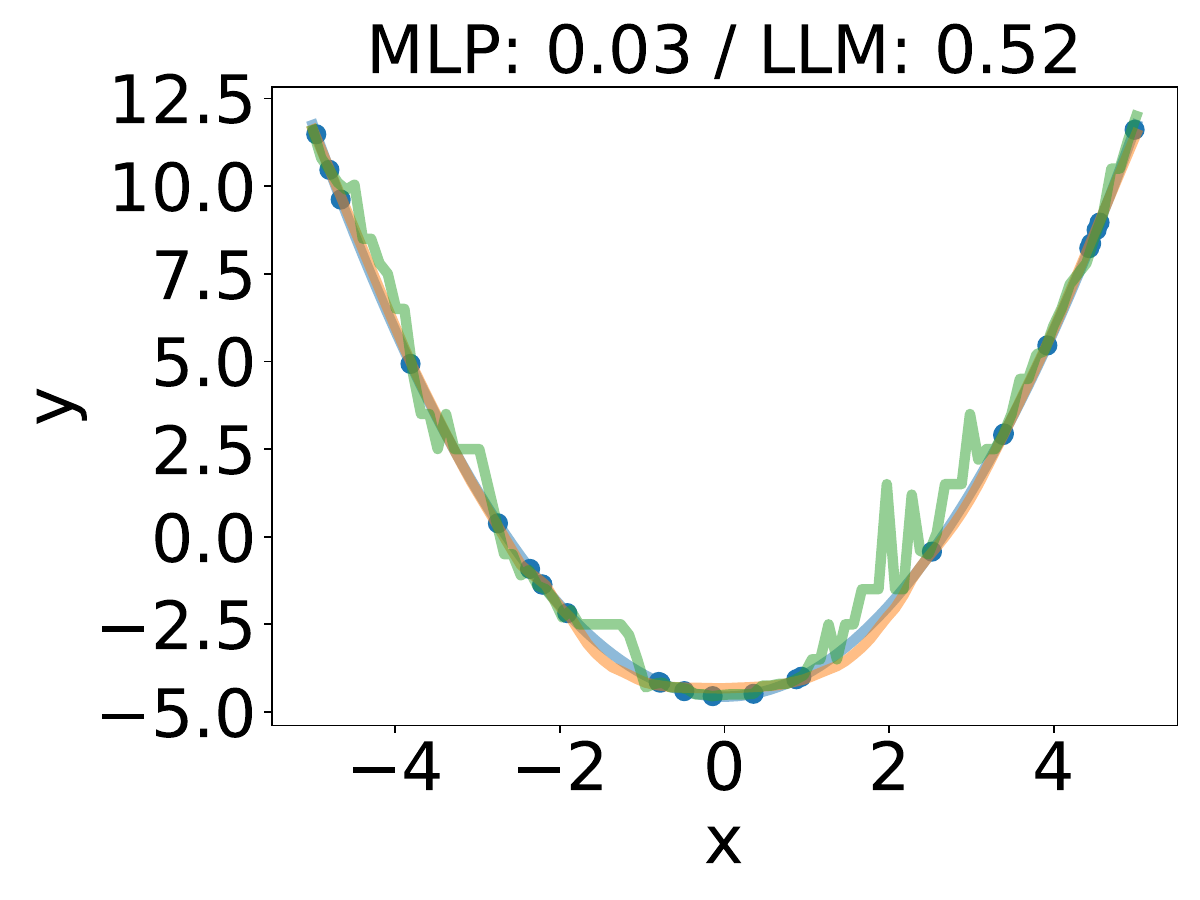}
        \caption{Quadratic}
    \end{subfigure}
    \hfill
    \begin{subfigure}[b]{\figsize}
        \centering
        \includegraphics[width=\textwidth]{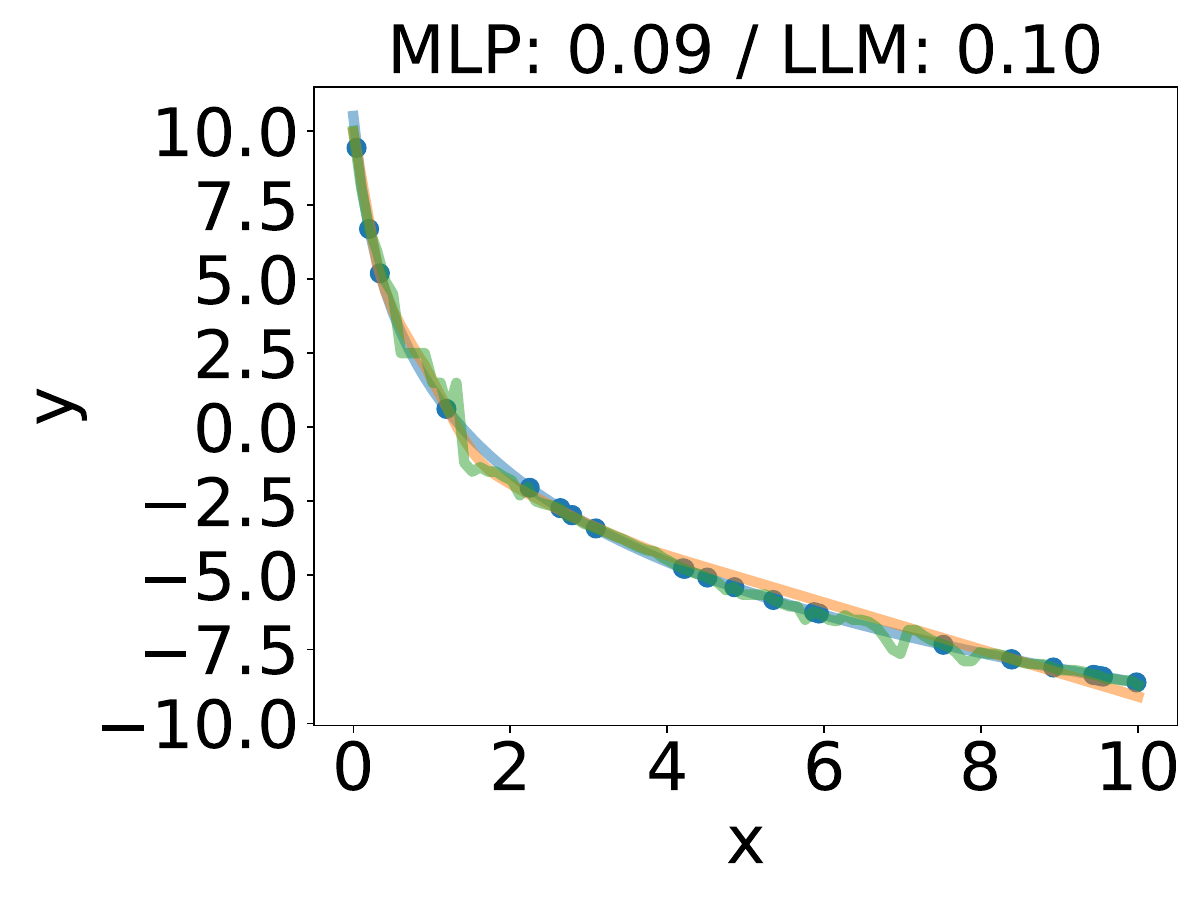}
        \caption{Logarithmic}
    \end{subfigure}
    \hfill
    \begin{subfigure}[b]{\figsize}
        \centering
        \includegraphics[width=\textwidth]{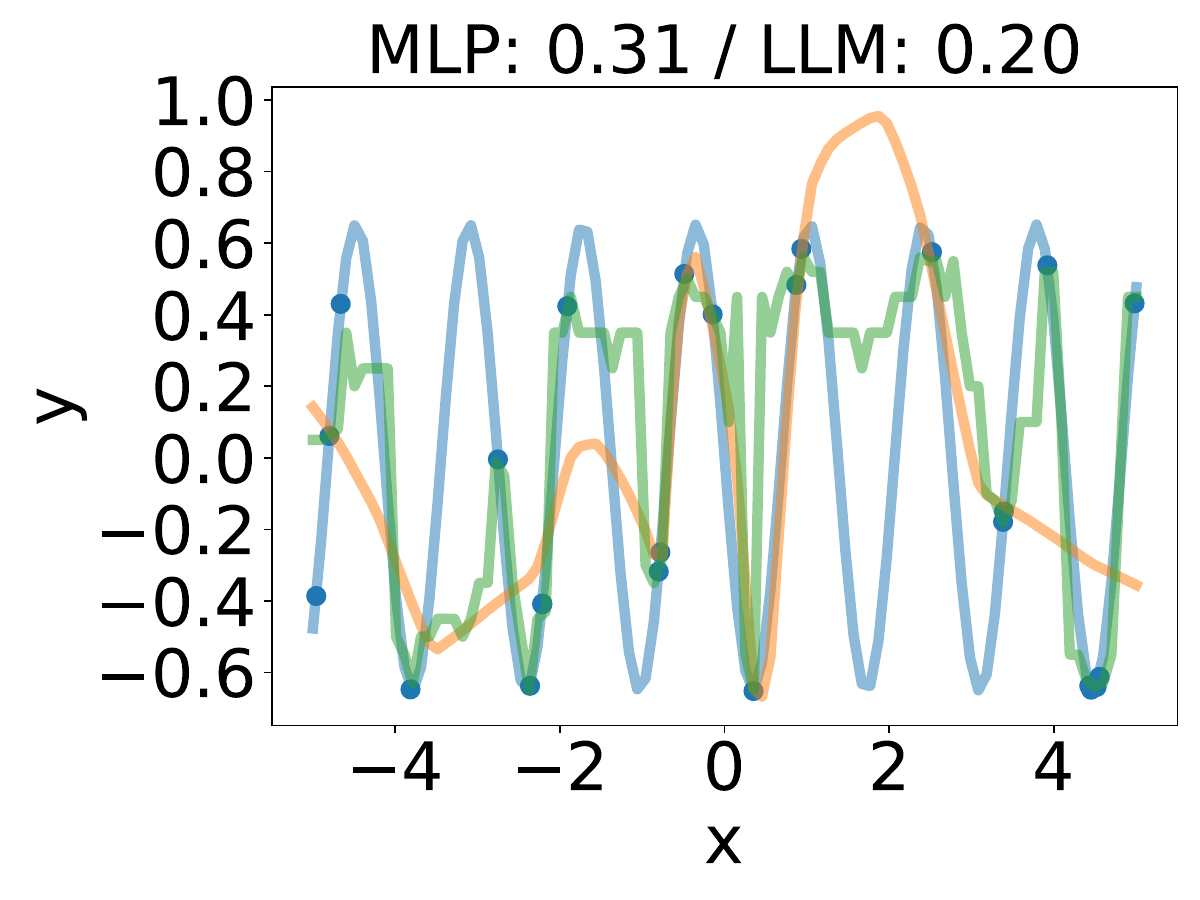}
        \caption{Sine}
        \label{fig:synthetic_functions_gpt4_noise_none_train_25.sine}
    \end{subfigure}
    \hfill
    \begin{subfigure}[b]{\figsize}
        \centering
        \includegraphics[width=\textwidth]{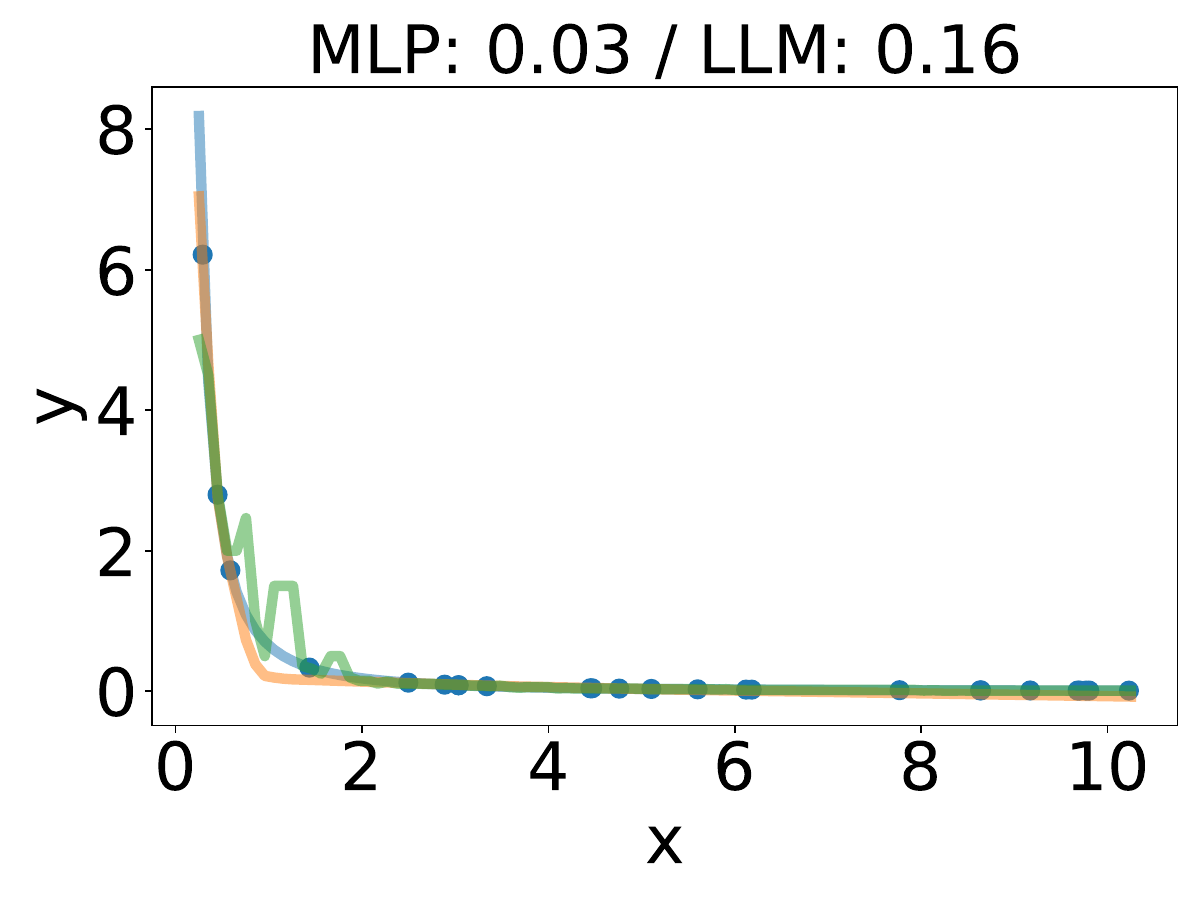}
        \caption{Power Law}
    \end{subfigure}

    \begin{subfigure}[b]{\figsize}
        \centering
        \includegraphics[width=\textwidth]{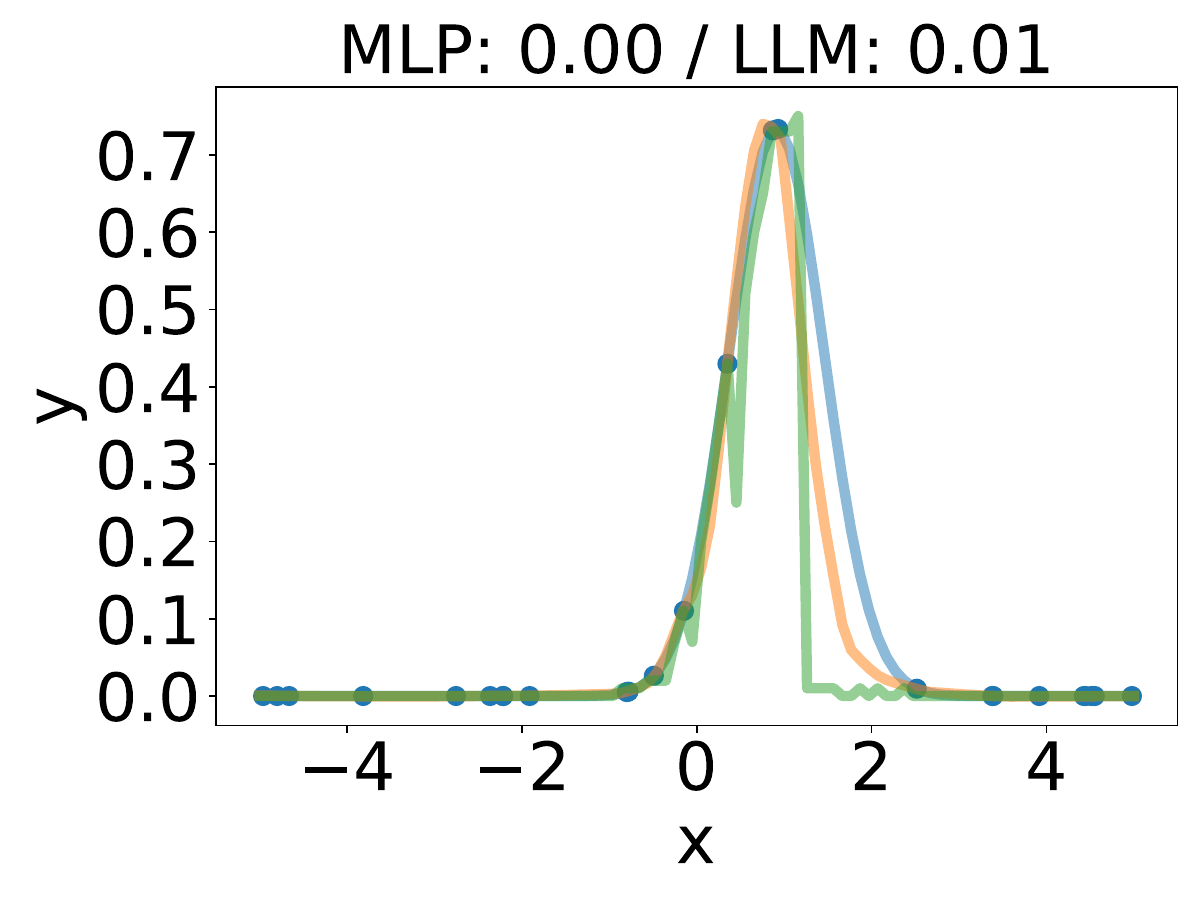}
        \caption{Gaussian}
    \end{subfigure}
    \hfill
    \begin{subfigure}[b]{\figsize}
        \centering
        \includegraphics[width=\textwidth]{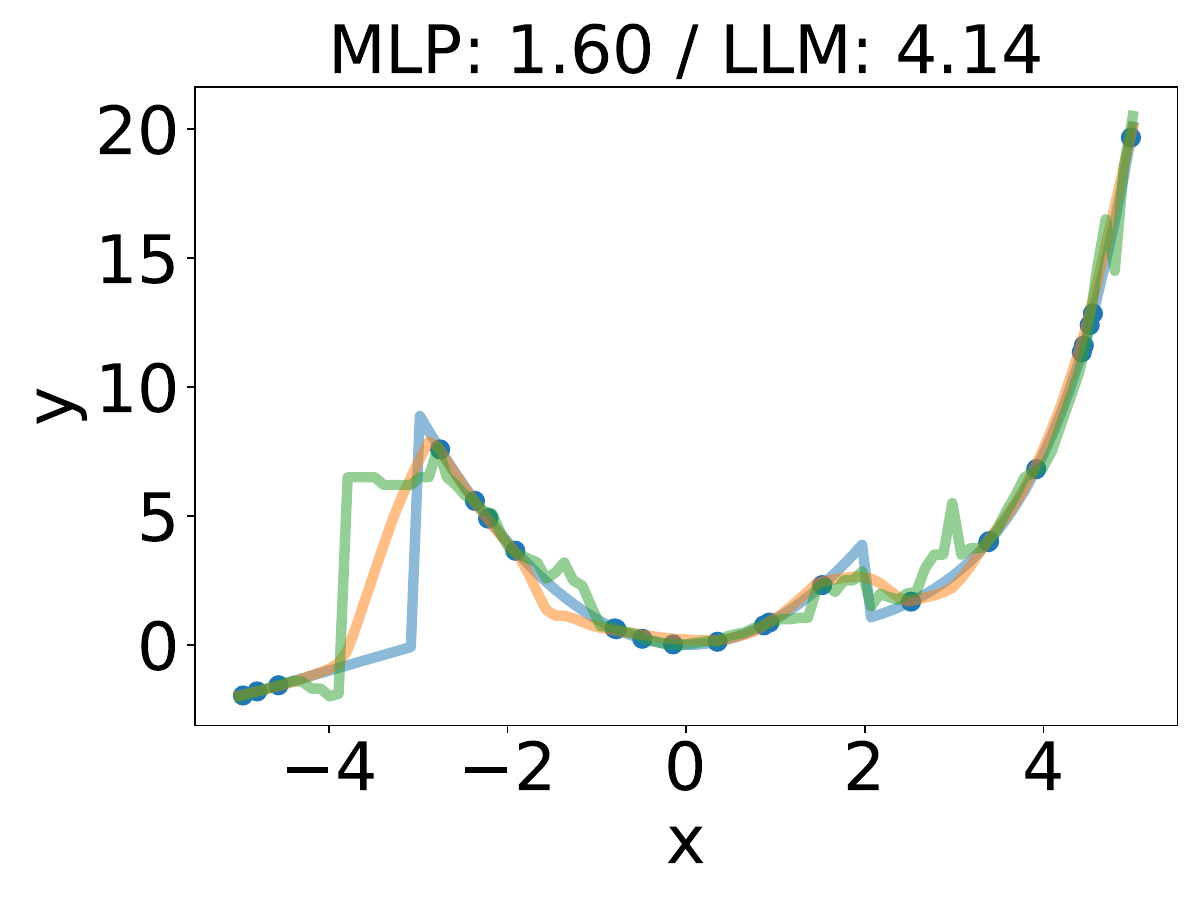}
        \caption{Piecewise}
    \end{subfigure}
    \hfill
    \begin{subfigure}[b]{\figsize}
        \centering
        \includegraphics[width=\textwidth]{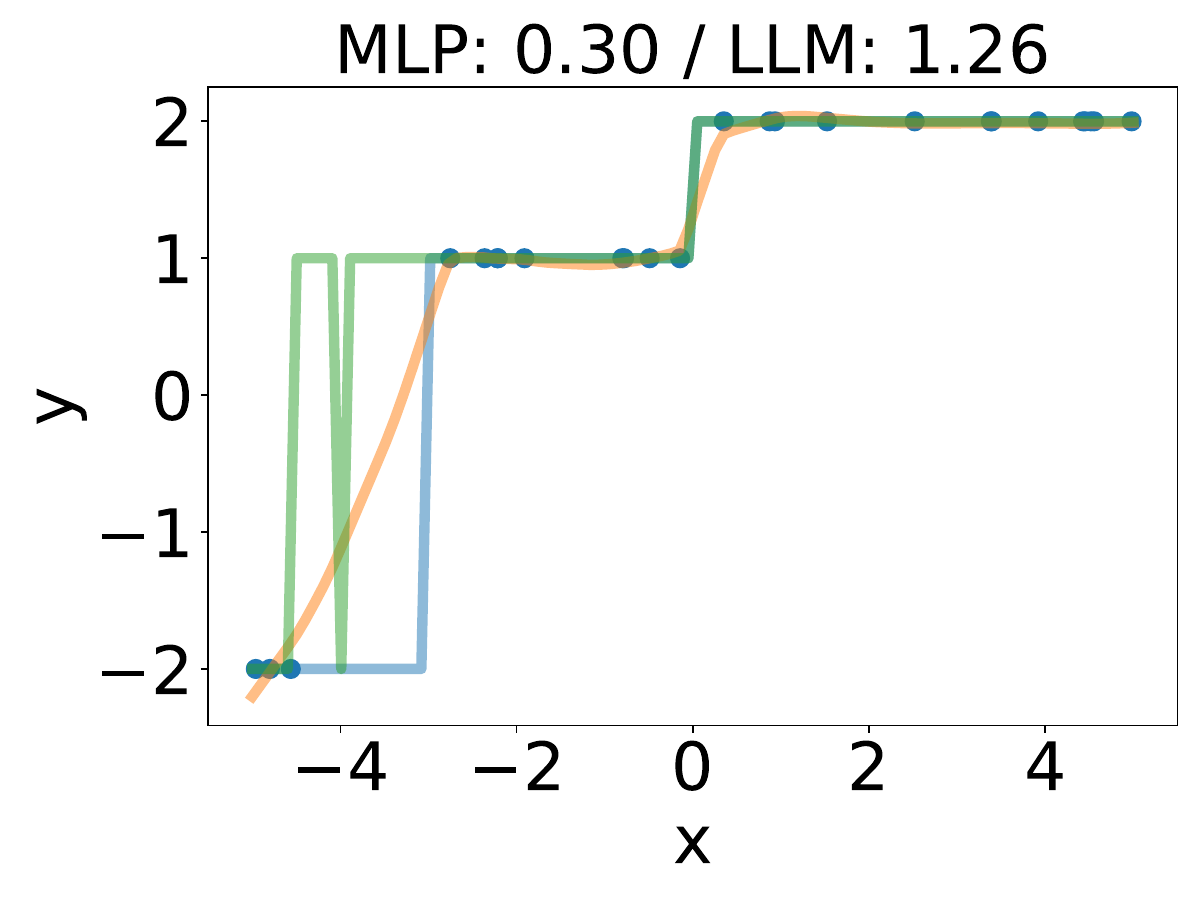}
        \caption{Step}
    \end{subfigure}
    \hfill
    \begin{subfigure}[b]{\figsize}
        \centering
        \includegraphics[width=\textwidth]{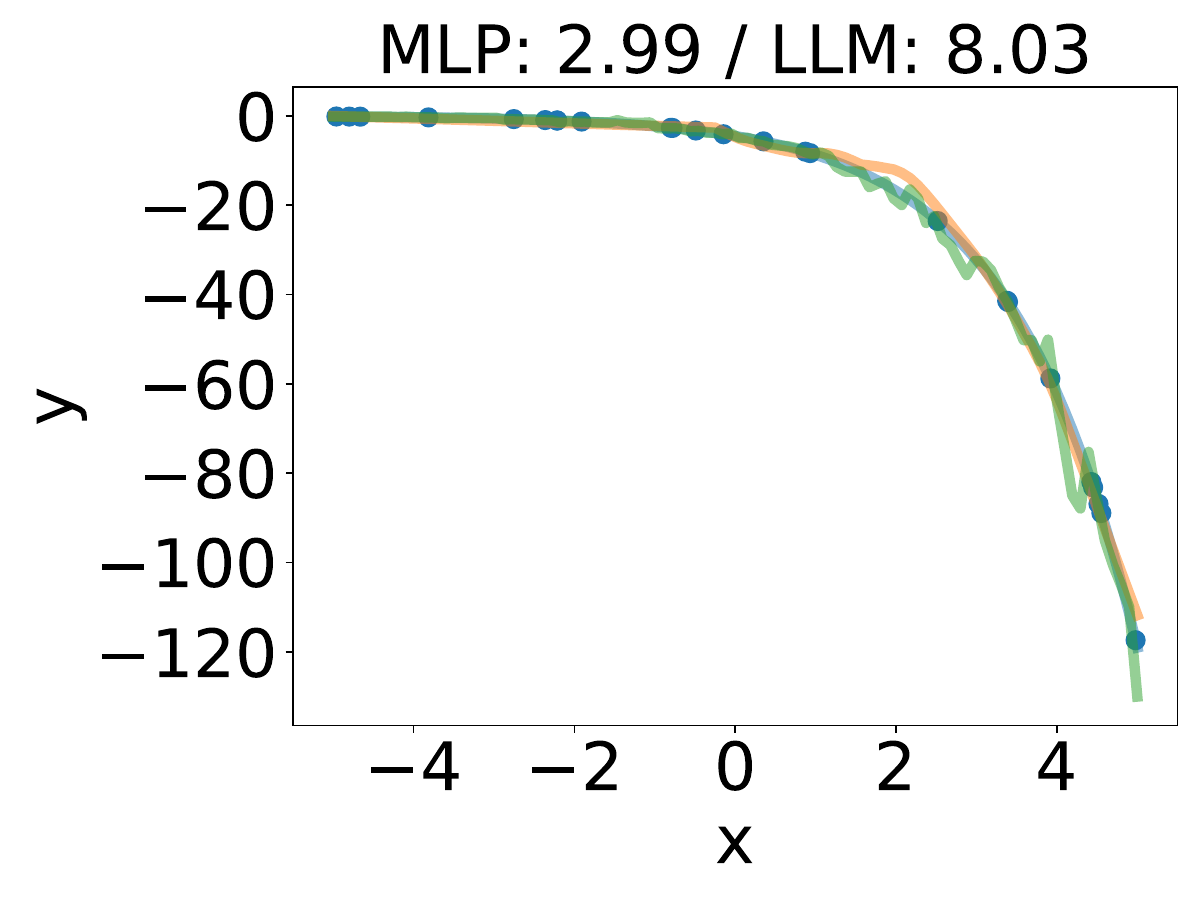}
        \caption{Exponential}
    \end{subfigure}
    \hfill
    \begin{subfigure}[b]{\figsize}
        \centering
        \includegraphics[width=\textwidth]{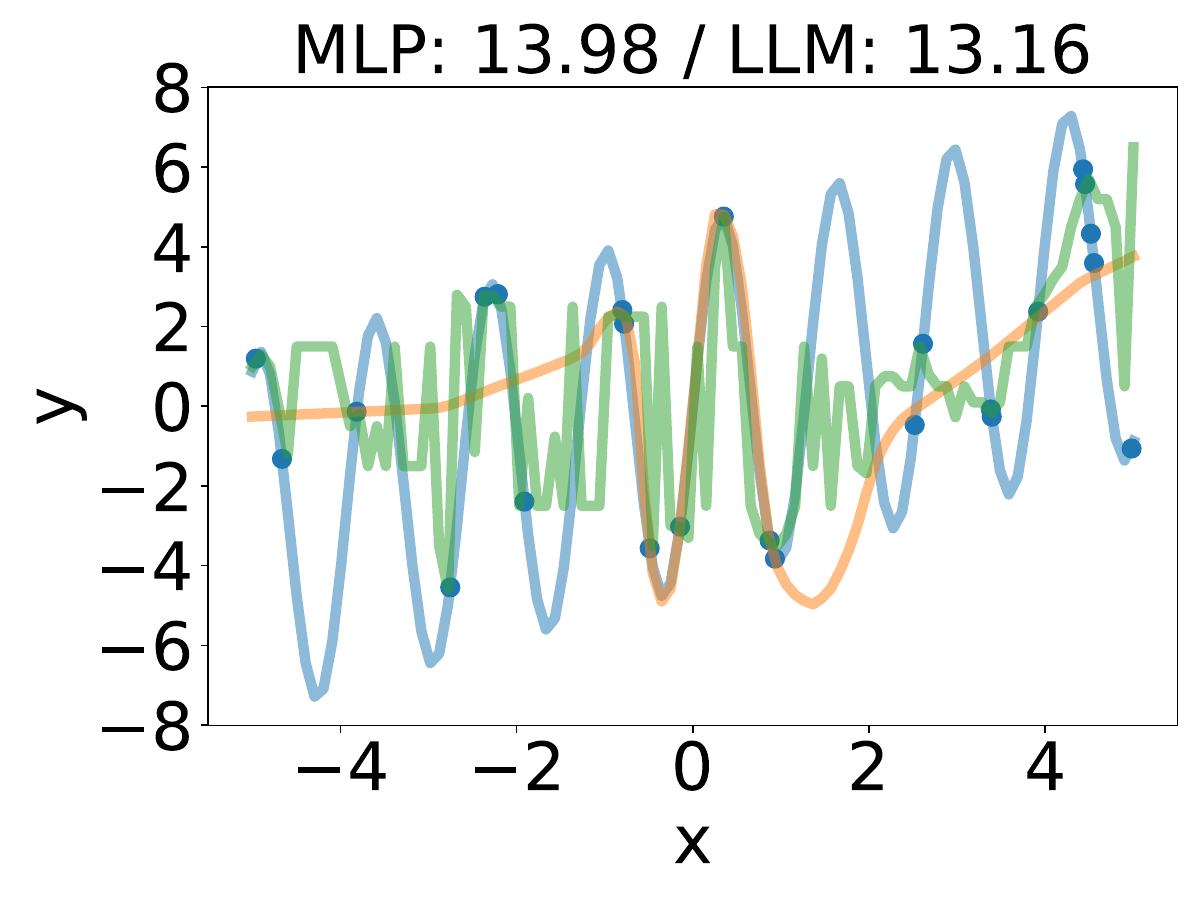}
        \caption{Periodic Linear}
        \label{fig:synthetic_functions_gpt4_noise_none_train_25.periodic_linear}
    \end{subfigure}

    \vspace{-2mm}
    \caption{Basic evaluations of function modeling using 25 training points, where we compare LLM performance (in particular GPT-4) with a 4-layer MLP with 64 hidden units. The MSE indicates direct prediction performance.}
    \label{fig:synthetic_functions_gpt4_noise_none_train_25}
\end{figure*}

\section{Experiments}

In this section, we evaluate the function modeling capabilities of LLMs in both synthetic and real-world tasks. We mainly focus on GPT-4 for our experiments, which was the most capable model during our evaluations, though our evaluation can fully be adapted to any other more recent models.

\subsection{Synthetic data}
\label{exp:synthetic_task}

\paragraph{Setup.} We first focus on evaluating LLMs' ability to understand patterns in raw data, where no domain knowledge is available. Here, we consider 10 types of commonly seen 1D functions. For each of these functions, we generate a total of 25 samples $\mathcal{D} = \{ x_i, y_i \}^{25}_{i=1}$. We ask the language model to make predictions for different query points $x'$ conditioned on the training examples.

\paragraph{Main results.} Unsurprisingly, we see that the performance of LLM (GPT-4 in this case) falls significantly below the performance of a simple MLP, where the gap in performance scales with the complexity of the dataset. While LLMs can model simple functions such as linear and quadratic functions accurately, they struggle in modeling more complex functions such as periodic functions (Fig.~\ref{fig:synthetic_functions_gpt4_noise_none_train_25.sine}) or composited functions (Fig.~\ref{fig:synthetic_functions_gpt4_noise_none_train_25.periodic_linear}). This highlights that language models struggle to model functions directly from raw data (except in the simplest of cases), which might be under-represented in the pretraining dataset~\citep{mccoy2023embers}.
Such inability may also be attributed to the tokenization process in language models, which can split numbers in a way such that they become ill-suited for further computation~\citep{spathis2024llmtokenization}.

\subsection{Income prediction}
\label{exp:income}

\paragraph{Setup.} We next consider a prediction task in social-economical study. The task here is to ask LLM to predict the income of an individual in the US given their personal information. The data $x$ in this task consists of 13 features $x = \{x_1, ..., x_{13}\}$ describing e.g. the age, occupation and education of the individual, and is taken from the UCI Adult dataset~\citep{misc_adult_2}. The binary target $y \in \{0, 1\}$ to predict is the income level of the individual (low v.s. high). Here the underlying function $f: \R^{13} \to \{0,1\}$ is multi-dimensional.

Since UCI Adult is a widely used dataset, a LLM may have seen this data during pre-training~\cite{orenproving}. To reduce the impact of potential test set memorization, which may lead to inaccurate evaluation of the language model's true function modeling capabilities, we modify the original dataset. Specifically, we (a) change the names of features (e.g. `education' $\rightarrow$ `degree'); (b) change feature values and scales by adding noise to the age feature and simplifying marital status to be binary; (c) replace features with equivalent counterparts e.g. `hours per week' $\rightarrow$ `hours per day'; (d) merge features such as merging capital gain/capital loss into capital net gain.
This results in a new dataset that is not directly seen by the language model.

A total number of $n=100$ samples are used for in-context learning.

\begin{table*}[ht!]
\caption{Income prediction: comparison of direction prediction performance of GPT-4 when w/ and w/o domain knowledge. The performance of a 4-layer MLP with 500 units trained with $n$ samples is also shown as reference.}
\vspace{-1mm}
\centering
\begin{tabular}{lcccc}
        \hline
            & \textbf{LLM w/o domain}  & \textbf{LLM w/ domain}  & \textbf{MLP} ($n=10^2$)  &  \textbf{MLP} ($n=10^4$) \\
        \hline
        test accuracy (\%) &  62.6 $\pm$ 0.61 &  79.4 $\pm$ 0.83 &  73.9 $\pm$ 0.45 & 82.0 $\pm$ 0.52  \\ 
        \hline
\end{tabular}
\label{tab:income}
\end{table*}

\paragraph{Main results.} In Table~\ref{tab:income}, we compare the predictive performance of LLM between the case when there is only raw data (denoted as LLM w/o domain) and the case when both the data and the domain information are provided (denoted as LLM w/ domain). We also highlight the performance of an MLP trained on the same data for reference. Our findings can be summarized as:
\begin{itemize}[leftmargin=*]
\item \emph{Raw data only:} The language model struggles to make accurate predictions in this case, as evident by the significant gap in performance as compared to the MLP. This highlights the difficulty in modeling complex multi-dimensional relationships just based on raw data using a language model.
These results are consistent with our findings on the synthetic dataset.
\item \emph{With additional domain knowledge:} The performance of the language model conditioned on the domain knowledge improves significantly, which is on par with an MLP trained on two orders of magnitude more data. 
\end{itemize}

The gap between the two cases highlights LLMs' efficacy in incorporating prior knowledge to update their understanding of the underlying function, compensating for their limitations in processing raw patterns. The results also underscore LLM's potential in small data regimes, where domain priors provide valuable compensation for limited data.

\paragraph{Further analysis.} In addition to comparing prediction accuracy, we further employ two evaluation methods: prediction interpretation and feature selection to gain further insight into the differences in the model's interpretation of the underlying function $f$ with and without domain knowledge.

\begin{table*}
\caption{Income prediction: Comparing the main prediction rules GPT-4 found when w and w/o domain knowledge. }
  \centering
\begin{tabularx}{\textwidth}{lXX}
    \hline
     & \qquad \qquad \textbf{Rules found w/o domain} & \qquad \qquad \textbf{Rules found w/ domain} \\
      \hline
\vspace{-0.25cm} \tt \small 1  &
\begin{spacing}{0.45}
\tt \small If feature 0 (age) and feature 2 (representative weight) are greater than 1, the sample is likely to be within class 1 (high income).
\end{spacing}
&
\begin{spacing}{0.45}
\tt \small People who have higher education (Masters, Doctorate, Bachelors) tend to have higher income. 
\end{spacing}
\\
    \hline
\vspace{-0.25cm} \tt \small 2  &
\begin{spacing}{0.45}
\tt \small Higher values of feature 2 (representative weight) often correspond to Class 1, while values around 0 seem more correlated with Class 0 (low income).
\end{spacing}
&
\begin{spacing}{0.45}
\tt \small Individuals with 'Married-civ-spouse' marital status are more correlated to higher income than individuals who are 'Never-married' or 'Divorced'.
\end{spacing}
\\
\hline
\vspace{-0.25cm} \tt \small 3  &
\begin{spacing}{0.45}
\tt \small If feature 8 (gender) is 0 and feature 0 (age) is below 0.5, the sample is  likely Class 1.
\end{spacing}
&
\begin{spacing}{0.45}
\tt \small Occupation such as 'Exec-managerial' and 'Prof-specialty' are better paid.
\end{spacing}
\\
\hline
  \end{tabularx}
\label{tab:rules-income}
\end{table*}

\begin{table*}[t]
\caption{Income prediction: Comparing the features selected by GPT-4 and that by state-of-the-art feature selection method. Accuracy is measured by an MLP trained using $10^4$ samples with the selected features on the test set. }
\vspace{-2mm}
\centering
\begin{tabularx}{\textwidth}{lXXX}
        \hline
          & \textbf{LLM w/o domain} & \textbf{LLM w/ domain} & \citep{yamada2020feature} \\
           &  ($n=10^2$) & ($n=10^2$) & ($n=10^4$) \\
        \hline
        top 5 features  & \tt \small \{gender, representative weight, age, {\color{green}hours per day}, ethnicity\} & \tt \small  \{{\color{green} degree, martial status},  occupation, {\color{green} hours per day, capital net gain}\}  &   \tt \small  \{{\color{green}degree, martial status,  age, hours per day, capital net gain}\} \\
        \hline
        test accuracy (\%) & 76.22 $\pm 0.56$ & 82.63 $\pm 0.40$ & 82.95 $\pm 0.38$ \\ 
        \hline
\end{tabularx}
\label{tab:fs}
\end{table*}

In \emph{prediction interpretation}, we ask the language model to output the rules used in prediction\footnote{\label{prediction_interp_eval_footnote}This is similar in spirit to interpretability derived from chain-of-thought traces~\citep{wei2022cot}. Note that these explanations could be misaligned with the actual rules used by the model~\citep{madsen2024self} due to well-known problems regarding language model hallucination~\citep{huang2023hallucinationsurvey}.}. These rules reflect the language model's understanding of the underlying functions. Table~\ref{tab:rules-income} summarizes the rules for cases with and without domain knowledge respectively. Comparing the two sets of rules, we discover that when domain knowledge is available, LLM tends to make use of robust features that align well with common sense, whereas when there is no domain knowledge, it relies more on spurious features that may only be predictive for the provided in-context examples. This difference again highlights that state-of-the-art LLMs such as GPT-4 can effectively utilize domain knowledge to improve their function modeling capabilities.

In \emph{feature selection}, we ask LLM to find the top-$k$ features $X' \subset X$ which together as a whole are most informative about the target $Y$:
\begin{equation}
    \max_{X': X' \subset X, |X'| = k} I(X'; Y)
\end{equation}
where $I(\cdot; \cdot)$ denotes mutual information. To select good features, one needs to understand both (a) the individual contribution of each feature to the underlying function and (b) how features interact with each other (e.g., synergy effects, redundancy, etc.). As such, this serves as a useful proxy for evaluating LLMs' function modeling capabilities. Table~\ref{tab:fs} summarizes the results. Leveraging its domain knowledge, GPT-4 is able to select a subset of features that closely matches the output of state-of-the-art feature selection methods~\citep{yamada2020feature}, while being two orders of magnitude more data efficient. In contrast, when relying solely on raw data, the model selects a poor set of features. These results highlight the impact of domain knowledge on LLM's function modeling process.

\subsection{CO$_2$ emission level modeling}
\label{exp:co2}

\begin{figure*}[t]
\centering
\hspace{-0.01\textwidth}
\begin{subfigure}{0.33\textwidth}
\centering
\includegraphics[width=1.0\linewidth]{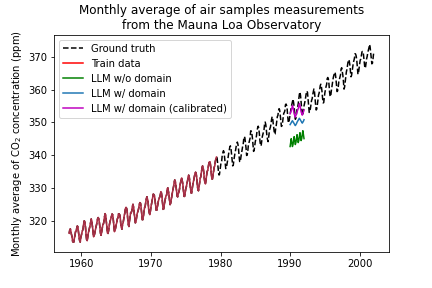}
\caption{\centering LLM prediction}
\label{fig:CO2_direct}
\end{subfigure}
\hspace{-0.01\textwidth}
\begin{subfigure}{0.33\textwidth}
\centering
\includegraphics[width=1.0\linewidth]{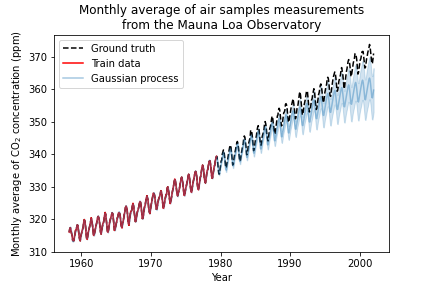}
\caption{\centering GP (expert-designed kernel)}
\label{fig:CO2_GP}
\end{subfigure}
\hspace{-0.01\textwidth}
\begin{subfigure}{0.33\textwidth}
\centering
\includegraphics[width=1.0\linewidth]{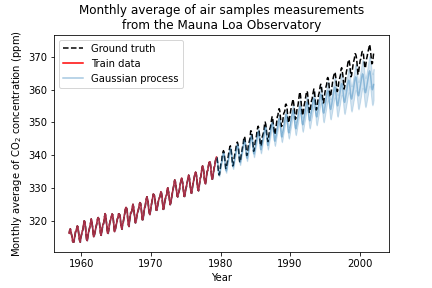}
\caption{\centering GP (LLM-designed kernel)}
\label{fig:CO2_GP_LLM_kernel}
\end{subfigure}
\caption{CO$_2$ level modeling: (a) Predictions made by GPT-4 with and without domain knowledge. (b-c) Predictions made by Gaussian processes with various kernels. The expert kernel is taken from \citep{williams2006gpbook}. }
\label{fig:CO2}
\vspace{-2mm}
\end{figure*}

\begin{table*}
\caption{CO$_2$ level modeling: Comparing the main prediction rules GPT-4 uses w/ and w/o domain knowledge. }
  \centering
\begin{tabularx}{\textwidth}{lXX}
    \hline
     & \qquad \qquad \textbf{Rules found w/o domain} & \qquad \qquad \textbf{Rules found w/ domain} \\
      \hline
\vspace{-0.25cm} \tt \small 1  &
\begin{spacing}{0.45}
\tt \small \underline{Increase Trend}: Although Y has oscillations, there is a general upward trend. As X increases, the Y values tend to increase overall.
\end{spacing}
&
\begin{spacing}{0.45}
{\tt \small \underline{Growing Trend}: There is a clear growth of  CO$_2$ concentration from 1958 to 1975, aligning with established knowledge that human activities (e.g. fossil fuel burning, deforestation) and industry contribute to rising CO$_2$ levels.}
\end{spacing}
\\
    \hline
\vspace{-0.35cm} \tt \small 2  &
\begin{spacing}{0.45}
\tt \small \underline{Periodic Oscillations in Y}: There seems to be a cyclical or periodic pattern in the Y values. For example, Y rises and falls several times, suggesting a wave-like behavior with peaks and troughs, even though X increases steadily. 
\end{spacing}
&
\begin{spacing}{0.45}
\tt \small \underline{Seasonality}: There's a seasonal pattern. CO$_2$ concentrations peak during early Northern Hemisphere spring due to reduced plant growth, and they reach a minimum during early fall when plant growth peaks.
\end{spacing}
\\
\hline
\vspace{-0.35cm} \tt \small 3  &
\begin{spacing}{0.45}
\tt \small \underline{Amplitude of Oscillations Grows Over Time}: The oscillations in Y seem to become more pronounced as X increases.
\end{spacing}
&
\begin{spacing}{0.45}
\tt \small \underline{Increasing Rate of Change}: Notably, the rate of CO$_2$ increase is accelerating. This corresponds with the 20th-century surge in industrial activity.
\end{spacing}
\\
\hline
  \end{tabularx}
\label{rules-income}
\vspace{-3mm}
\end{table*}

\paragraph{Setup.} We further consider a function modeling task in climate science. In this task, we would like to ask LLM to predict the CO$_2$ concentration level $y \in \R$ given the time $x \in (1975, 2000)$. The data is collected in the Mauna Loa Observatory \cite{co2_dataset}. The underlying function $f: \R \to \R$ is one-dimensional.

In order to reduce the impact of potential test set memorization~\cite{orenproving}, we similarly employ techniques to update the dataset, including (a) hiding the information about the exact observatory that this data was collected from; (b) add random Gaussian noise $\epsilon \sim \mathcal{N}(\epsilon; 1, 10^{-2})$ to the measurements; and (c) shift the data by 1 unit, creating an unseen dataset.

We use data up to year 1980 as our training set (used via in-context learning), and use the data between year 1990 and year 1992 as our test set.

\paragraph{Main results.} In Fig.~\ref{fig:CO2_direct}, we compare the predictive performance of LLM on the test set between cases when there is only raw data (denoted as LLM w/o domain) and the case when both the data and the domain information are provided (denoted as LLM w/ domain). For reference, we also show the performance of a Gaussian process (GP) trained with the same data in Fig.~\ref{fig:CO2_GP} with an expert-chosen kernel taken from~\citep{williams2006gpbook}. Consistent with prior results, domain knowledge plays a critical role in transforming a language model into an accurate functional approximator. To summarize:
\begin{itemize}[leftmargin=*]
    \item \emph{Raw data only:} When provided only with raw data, the model struggles to correctly model the concentration level. Specifically, it not only underestimates the CO$_2$ concentration level but also incorrectly models the frequency of the seasonal pattern, significantly underperforming compared to a GP trained on the same data. This result aligns with our earlier findings in Section~\ref{exp:income}, where we observed that the language model struggles to make accurate predictions for a similar 1D function (see Fig.~\ref{fig:synthetic_functions_gpt4_noise_none_train_25.periodic_linear}).
    \item \emph{With additional domain knowledge:} We see a significant boost in model performance with the inclusion of domain knowledge. Both the CO$_2$ concentration level and the seasonal period exhibit significant improvements in comparison to just raw data. Notably, predictions made by the language model in this case are comparable to an expert-designed GP (shown in Fig.~\ref{fig:CO2_GP}).
\end{itemize}

The gap between the two cases again highlights the critical role of the prior in the function modeling capabilities of LLMs, resulting in predictions that are comparable to, or even surpass, those of an expert-designed Gaussian process model.

\paragraph{Further analysis.}
Similar to the evaluation in Section~\ref{exp:co2}, we use prediction interpretation and kernel selection to gain better insights into the language model's function modeling process.

In \emph{prediction interpretation}, we ask LLM to verbalize the rules it used for prediction\footnote{See caveat presented in footnote~\ref{prediction_interp_eval_footnote}.}. Table~\ref{rules-income} highlights the rules that the language model assumes to be using. For this task, both cases correctly recognize the two main types of patterns in the data i.e., the overall increasing trend and the seasonality. However, there is a significant improvement in modeling capability with domain knowledge. For instance, when modeling seasonality, domain knowledge helps pinpoint the exact peak and the duration of the cycle, making it more precise than the case with raw data alone. Crucially, with domain knowledge, the LLM can also detect patterns not directly apparent in the data, such as the increasing rate of change. When explicitly prompting the language model to calibrate its prediction to account for this pattern, it produces estimates that outperform an expert-designed GP. These results clearly demonstrate that LLMs possess a strong prior of the underlying function, and can effectively leverage this prior to improve   function modeling.

In \emph{kernel design}, we further test the language model's understanding of the underlying function by asking the language model to design the Gaussian Process (GP) kernel, which reflects the assumptions about the underlying function being modeled\footnote{A kernel measures the similarity between two inputs in functional space, thereby implicitly modeling the function. }~\citep{williams2006gpbook}. We evaluate whether LLMs like GPT-4 can come up with good kernels by incorporating domain knowledge. 
Given data and domain specification, the language model suggests the following kernel that achieves a comparable performance to an expert-designed kernel: 
\begin{equation}
    k_{llm}(t, t') = \sum^4_{l=1} k_l(t, t') 
\end{equation}
where $k_1, k_2, k_3, k_4$ are the RBF kernel, the exponential sine squared kernel, the rational quadratic kernel and the white noise kernel, respectively. These kernels correspond to (1) the long-term trend of CO$_2$ emission; (2) the yearly cyclical pattern; (3) the short-term fluctuations; and (4) observation noise and unmodeled factors respectively. The choice of these kernels by GPT-4 reflects a strong understanding of the underlying function when provided with both raw data and domain knowledge.

\section{Conclusion}

In this work, we introduced a novel evaluation framework to systematically assess the function modeling capabilities of Large Language Models (LLMs). By disentangling their ability to understand raw data patterns from their ability to leverage prior knowledge, we identified both strengths and weaknesses in LLMs for function modeling tasks, namely (a) they struggle to understand functions based on just raw data, except for the simplest cases,  and (b) their true strength lies in incorporating domain knowledge. Our research provides a foundation for the reliable and effective applications of LLMs in real-world prediction tasks.

Our research suggests that future advancements in LLMs may benefit from explicitly enhancing their ability to understand raw data patterns during pretraining, which would significantly expand their applicability to numerical prediction tasks. Additionally, the development of powerful multimodal models that can simultaneously process numerical data and understand contextual information presents a promising research direction for future.

\section*{Limitation}

While our evaluation techniques are broadly applicable to any language model, they are predominantly based on GPT-4 due to space constraints, as it was the most powerful model at the time of writing. Additionally, the experimental results may vary over time due to model updates. We therefore advise readers to interpret our findings cautiously, though our evaluation method can be fully reused for assessing future models.

Furthermore, our current evaluation focuses on the case of in-context learning, and is aligned with prior work in this space. We consider the evaluation of finetuned models to be an exciting avenue for the future.

Finally, as noted in the literature, LLM outputs are highly dependent on input prompts~\citep{lester2021power}. Therefore, we assume the exact results can vary to a moderate extent based on the prompting technique used. However, we expect our findings to generalize across these different prompting techniques.

\section*{Acknowledgements}

The authors would like to acknowledge Azure credits provided via \textit{Microsoft Accelerate} program used for the project.

\bibliography{refs}

\end{document}